\documentclass[twocolumn]{autart}    

\usepackage{graphicx}      

\usepackage[round,authoryear]{natbib}

\usepackage{amsmath}      
\usepackage{mathrsfs}
\usepackage{algpseudocode} 
\usepackage{amssymb} 
\usepackage{algorithm}
\usepackage{mathtools} 
\usepackage{booktabs} 
\usepackage{url}

\allowdisplaybreaks[4]
\begin{document}

\begin{frontmatter}

\title{Momentum LMS Theory beyond Stationarity: Stability, Tracking, and Regret\thanksref{footnoteinfo}}


\thanks[footnoteinfo]{This paper was supported by the National Key Research and Development Program under Grant No. 2024YFC3307200 and the National Natural Science Foundation of China under Grant No. 12288201.}
\thanks[footnoteinfo2]{Corresponding author: Lei Guo.}

\author[a,b]{Yifei Jin}\ead{jinyifei@amss.ac.cn}
\author[b,c]{Xin Zheng}\ead{zhengxin2021@amss.ac.cn}
\author[b,c]{Lei Guo\thanksref{footnoteinfo2}}\ead{lguo@amss.ac.cn}

\address[a]{School of Advanced Interdisciplinary Sciences, University of Chinese Academy of Sciences, Beijing 101408, China.}

\address[b]{State Key Laboratory of Mathematical Sciences, Academy of Mathematics and Systems Science, Chinese Academy of Sciences, Beijing 100190, China.}

\address[c]{School of Mathematical Sciences, University of Chinese Academy of Sciences, Beijing 100049, China.}


\begin{abstract}                          
In large-scale data processing scenarios, data often arrive in sequential streams generated by complex systems that exhibit drifting distributions and time-varying system parameters. This nonstationarity challenges theoretical analysis, as it violates classical assumptions of i.i.d. (independent and identically distributed) samples, necessitating algorithms capable of real-time updates without expensive retraining. An effective approach should process each sample in a single pass, while maintaining computational and memory complexities independent of the data stream length. Motivated by these challenges, this paper investigates the Momentum Least Mean Squares (MLMS) algorithm as an adaptive identification tool, leveraging its computational simplicity and online processing capabilities. Theoretically, we derive tracking performance and regret bounds for the MLMS in time-varying stochastic linear systems under various practical conditions. Unlike classical LMS, whose stability can be characterized by first-order random vector difference equations, MLMS introduces an additional dynamical state due to momentum, leading to second-order time-varying random vector difference equations whose stability analysis hinges on more complicated products of random matrices, which poses a substantially challenging problem to resolve. Experiments on synthetic and real-world data streams demonstrate that MLMS achieves rapid adaptation and robust tracking, in agreement with our theoretical results especially in nonstationary settings, highlighting its promise for modern streaming and online learning applications.
\end{abstract}
\end{frontmatter}

\section{Introduction}
Over the past few decades, a wide range of machine learning methods has achieved substantial success under the classical offline-training–online-generalization paradigm, 
where models are trained on fully available datasets prior to deployment and subsequently used for prediction or other tasks in operation \citep{goodfellow2016deep}. However, this paradigm is increasingly inadequate for the growing number of applications driven by streaming data, such as online recommendation systems~\citep{egg2021online,chang2017streaming}, credit scoring~\citep{west2000neural}, sensor networks~\citep{gama2007learning}, and reinforcement learning in artificial intelligence~\citep{sutton1998reinforcement}, where data arrive sequentially in real time and need to be processed immediately to enable timely decision-making.
Therefore, the paradigm of storing the entire dataset before processing it becomes increasingly infeasible, as the data volume may be extremely large while computational resources are often limited in practice, necessitating methods that can operate effectively under streaming conditions.

To tackle these challenges, online learning has emerged as a key paradigm. Such methods update models on the fly with each arriving sample, eliminating the need to revisit past data, reducing storage requirements, and enabling single-pass processing over potentially unbounded streams. Early online learning methods such as the Perceptron~\citep{rosenblatt1958perceptron} laid important groundwork for the development of more advanced and 
task-specific online algorithms. For instance, \citet{zinkevich2003online} studied sequential convex decision-making and proposed projected online gradient descent with regret guarantees, while \citet{zhang2018adaptive} extended this framework to nonstationary settings through Adaptive Learning for Dynamic Environment (Ader), a meta-algorithm that aggregates multiple projected gradient descent experts to achieve near-optimal dynamic regret. Online Area Under Curve (AUC) Maximization~\citep{zhao2011online} targets imbalanced binary classification by optimizing AUC in a streaming setting via gradient-based updates and reservoir sampling. For supervised learning under unknown distribution drift, \citet{ mazzetto2024improved} focused on estimating a sequence of drifting discrete distributions from independent samples by designing distribution-dependent, adaptive windowing schemes. In streaming time-series forecasting, Online ensembling Network (OneNet, \cite{wen2023onenet}) addressed concept drift by maintaining an online ensemble of complementary neural forecasters whose weights are continuously updated. 

A fundamental challenge in many online learning scenarios is the inherent non-stationarity of streaming data. The underlying data distributions may drift over time, and system parameters may evolve in response to external perturbations (see, e.g., \cite{ljung1990adaptation}). These characteristics highlight the necessity of a comprehensive study of adaptive learning algorithms grounded in gradient- and Newton-based optimization methods, which form the foundation of modern machine learning, adaptive signal processing, system identification, and feedback control. For stochastic regression models driven by general nonstationary signals and time-varying parameters, a unified framework for the stability and performance analysis of three fundamental adaptive tracking schemes (the Kalman filter, least mean squares (LMS), and recursive least squares (RLS)) was provided in \citep{guo1994stability, guo1995exponential, guo1995performance}.

Among these algorithms, the LMS-type adaptive method stands out for its structural simplicity and minimal computational cost, features that have enabled its widespread practical applications. Consequently, extensive theoretical investigations of LMS-type algorithms have been carried out over the past half-century (see, e.g., \cite{widrow1985adaptive, solo1994adaptive, Macchi1995AdaptiveProcessing, hassibi1995lms, kushner1995analysis, haykin2002adaptive, sergio2002adaptive, guo2020time}), following the pioneering work of \cite{WidrowHoff1976}. This line of research culminated in a most comprehensive analysis of the standard LMS algorithm under general nonstationary signal conditions presented in \cite{GuoLjungPriouret1997}.

\begin{figure}
\begin{center}
\includegraphics[width=1\linewidth]{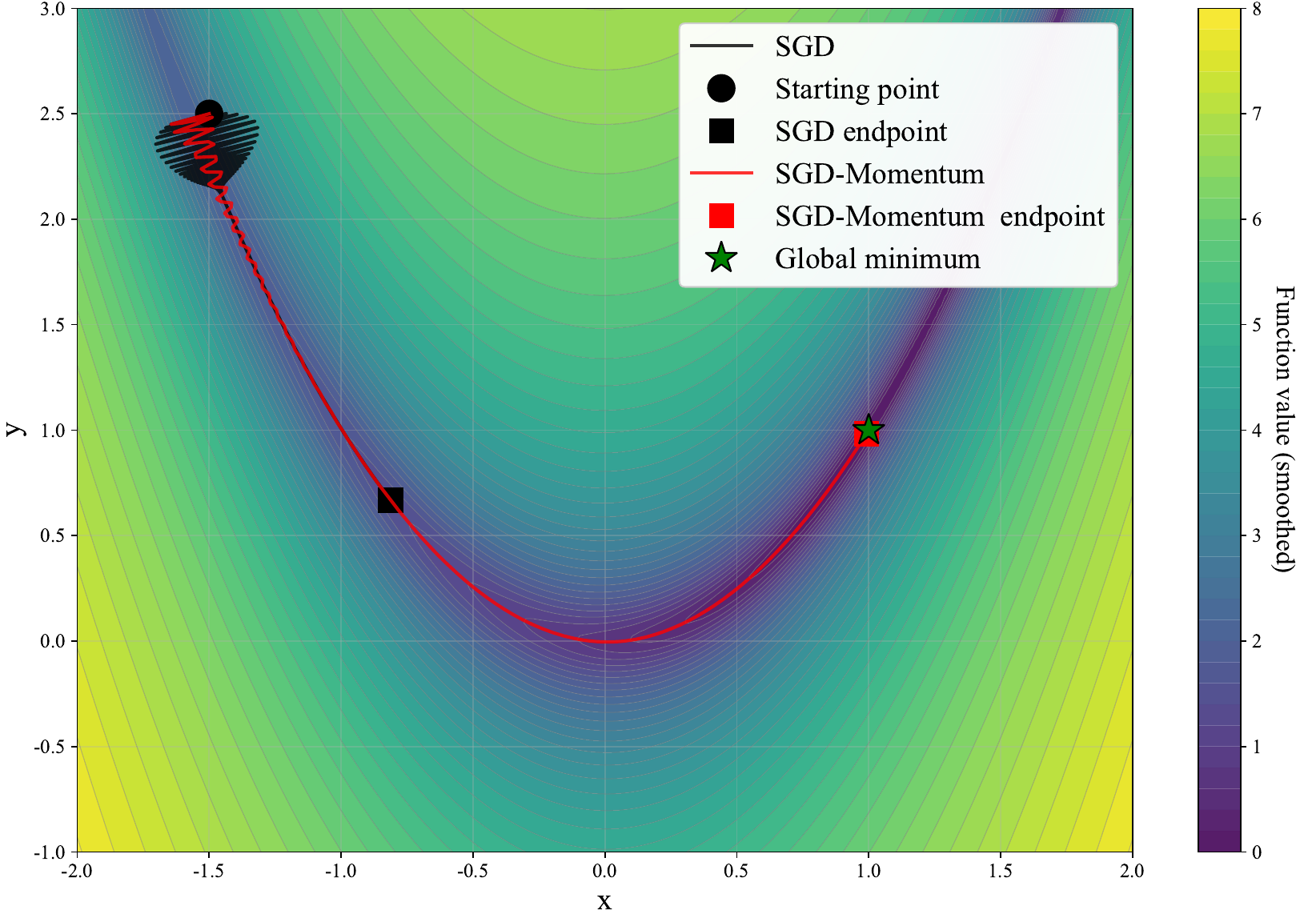}    
    \caption{Trajectories of SGD and SGD-Momentum on the Rosenbrock function. The green star marks the global optimum $(1,1)$.
    Starting from $(-1.5,\,2.5)$ with the same step-size, SGD (black) zigzags across the
    curved, ill-conditioned valley, while SGD with momentum (red) builds a velocity that
    damps high-curvature oscillations and accelerates motion along the valley floor, producing a smoother path,
    fewer iterations, and a lower final objective value.}
    \label{fig1}                               
\end{center}                                 
\end{figure}

On the other hand, as illustrated in Fig.~\ref{fig1}, the traditional Stochastic Gradient Descent (SGD) algorithm (equivalent to the classical LMS recursion for a linear regression model with a quadratic cost) can be enhanced by incorporating a Polyak-type momentum term (see, e.g., \cite{polyak1964some}). As noted by \cite{Widrowbackpropagation}, ``The momentum technique low-pass filters the weight updates and thereby tends to resist erratic weight changes caused either by gradient noise or high spatial frequencies in the MSE surface.'' Although Momentum Least Mean Squares (MLMS) algorithms have been studied previously, the majority of existing work remains largely empirical in nature \citep{dahanayake1993derivation, CHAUDHARY2021412}. Meanwhile, the existing theoretical analyses typically rely on assumptions of time-invariant linear systems driven by stationary inputs that are frequently violated in practical, time-varying environments \citep{tugay1989properties, ting2000tracking, chhetri2006acoustic, roy2002analysis, shynk1988lms}.
  
To solve the aforementioned challenging issues, we consider the stability analysis of the MLMS algorithm under general data conditions. Specifically, the main contributions of this paper are summarized as follows:
\begin{itemize}
\item First, we provide a theoretical guarantee of MLMS in streaming environments with general data setting which does not rely on stationarity or independence data assumptions. 
A central technical contribution is to overcome the momentum-induced analytical challenge by augmenting the state to obtain a higher dimensional first order random difference equation, which is fundamentally different from that encountered in the analysis of LMS, since one needs to analyze more complicated product of random matrices than that associated with LMS.
When the underlying parameter is time-invariant, the established tracking guarantees naturally yield learning and, subsequently, generalization guarantees.

\item Second, we characterize the prediction performance of MLMS by deriving asymptotic upper bounds on the prediction regret, without requiring any excitation conditions. We further show that LMS-type algorithms, although originally developed for linear time-varying systems, could be effectively extended to a broad class of nonlinear models. This may provide an explanation for their widely observed empirical success in nonlinear applications.

\item Finally, we perform empirical evaluations on both synthetic and real-world data streams. The results demonstrate that MLMS achieves faster adaptation and higher predictive accuracy than the commonly used adaptive optimizers and empirically validate our theoretical analysis in nonstationary environments.
\end{itemize}

This paper is organized as follows. Section~\ref{sec:algo} presents the problem formulation and basic assumptions. Section~\ref{hello} introduces the MLMS algorithm for estimating time-varying parameters. Section~\ref{Results} states the main theoretical results. Section~\ref{formal} presents the formal linear representation, which transforms a broad class of nonlinear systems into linear time-varying systems. Section~\ref{experiments} reports simulations on both synthetic and real-world data streams. Finally, Section~\ref{conclusion} concludes the paper.
\section{Problem formulation}\label{sec:algo}
Consider the following time-varying linear regression model:
\begin{equation} \label{timevarying1}
    y_{k+1} = \varphi_k^{\top} \theta_k + v_{k+1}, 
    \quad k = 0, 1, 2, \ldots,
\end{equation}
where $\theta_k \in \mathbb{R}^m\,(m \ge 1)$ denotes the unknown time-varying parameter vector to be estimated at time $k$. The observation $y_{k+1} \in \mathbb{R}$ denotes the system output. The regressor $\varphi_k \in \mathbb{R}^m$ represents the stochastic input vector. The random variable $v_{k+1} \in \mathbb{R}$ corresponds to the noise.
\subsection{Notations and assumptions}
To facilitate our analysis, we introduce the following notations and assumptions.

\textbf{Notations.} For any random matrix $M \in \mathbb{R}^{m \times n}$, its Euclidean norm is defined as its maximum singular value, i.e., $\|M\|=\left\{\lambda_{\max }\left(M^\top M \right)\right\}^{\frac{1}{2}}$, and its $L_p$-norm is defined as $\|M\|_{p}=\left\{\mathbb{E}\left[\|M\|^p\right]\right\}^{\frac{1}{p}}$, where $\mathbb{E}[\cdot]$ denotes the expectation operator, $p \geq 1$. The maximum and minimum eigenvalues are denoted by $\lambda_{\max }\{M\}$ and $\lambda_{\min }\{M\}$ respectively. Moreover, $\left\{\mathcal{F}_k, k \geq 0\right\}$ is the sequence of $\sigma$-algebra together with that of conditional mathematical expectation operator $\mathbb{E}\left[\cdot \mid \mathcal{F}_k\right]$. A stochastic sequence $\{x_k,\mathcal{F}_k\}$ is said to be adapted if $x_k$ is $\mathcal{F}_k$-measurable for all $k\ge0$.  The floor function $\lfloor \cdot \rfloor$ maps a real number to the greatest integer less than or equal to it, i.e.,
$
\lfloor x \rfloor = \max\{k \in \mathbb{Z} \mid k \le x\}, x \in \mathbb{R}.
$
The ceiling function $\lceil \cdot \rceil$ maps a real number to the smallest integer greater than or equal to it, i.e.,
$
\lceil x \rceil = \min\{k \in \mathbb{Z} \mid k \ge x\}, x \in \mathbb{R}.
$

We also need the following assumptions:
\begin{assum}\label{condition2}
The stochastic regressor $\left\{\varphi_k, \mathcal{F}_k\right\}$ is an adapted sequence, where $\left\{\mathcal{F}_k, k \geq 0\right\}$ is a nondecreasing sequence of $\sigma$-algebras, and there exist a constant $\alpha>0$ and an integer $h>0$ such that for any $k\geq 0$
\begin{equation}\label{conditional pe}
\mathbb{E}\left[\left.\sum_{i=kh+1}^{(k+1)h} \frac{\varphi_i \varphi_i^\top}{\delta + \|\varphi_i\|^2} \right| \mathscr{F}_{kh}\right] \geq \alpha I > 0, \quad \text{a.s.}
\end{equation}
\end{assum}
\begin{rem}
It should be noted that this assumption was first introduced in \cite{guo2002estimating} and was shown to be a necessary condition for the stability of tracking algorithms under certain circumstances (see \cite{guo1994stability}). 
Clearly, Assumption~\ref{condition2} is more general than the commonly used deterministic
persistence of excitation (PE) condition: there exist a positive integer $N$ and
constants $0 < a \le b < \infty$ such that 
\begin{equation}\label{eufhu}
a I \le \sum_{k=n}^{n+N} \varphi_k \varphi_k^{\top} \le b I, \qquad \forall n \ge 0.
\end{equation}
Unfortunately, this deterministic condition is not suitable for stochastic or random data. In fact, \eqref{eufhu} may exclude useful unbounded random signals (e.g., Gaussian signals), and even appears to be hardly verifiable for bounded i.i.d. signals. Moreover, 
unlike the above deterministic PE condition \eqref{eufhu}, Assumption~\ref{condition2} does not impose independence or stationarity assumptions on the sequences $\{\varphi_k\}$ and may include signals generated from stochastic systems with feedback, which are clearly not i.i.d. signals.
\end{rem}
\section{The MLMS Algorithm}\label{hello}
To improve the adaptation of LMS in nonstationary environments, the MLMS algorithm augments LMS with a Polyak heavy-ball–type momentum term with an exponentially decaying memory of past updates, as detailed in the algorithm below.
\begin{algorithm}[htbp]
\caption{MLMS Algorithm}
\label{alg:mnlms}
\begin{algorithmic}[1]
    \Require Step-size $\mu \in (0,1)$;
             regularization constant $\delta > 0$; momentum coefficient \(\beta\) satisfying \(0\le\beta\le C_\beta\mu<1\), 
              for some $C_\beta>0$.
    \Ensure Predictions $\{\hat{y}_{k+1}\}$; parameter estimates $\{\hat{\theta}_k\}$.
    \State Initialize $\hat{\theta}_{-1} = \hat{\theta}_{0} = \mathbf{0}$.
    \For{$k = 0,1,2,\ldots$}
        \State Compute adaptive step-size:
        \[
            \alpha_k = \frac{\mu}{\delta + \|\varphi_k\|^2}.
        \]
        \State Predict output:
        \[
            \hat{y}_{k+1} = \varphi_k^{\top}\hat{\theta}_k.
        \]
        \State Calculate prediction error:
        \[
            e_k = y_{k+1} - \hat{y}_{k+1}.
        \]
        \State Update parameters with momentum:
\begin{equation}\label{eq:hfwhif}
            \hat{\theta}_{k+1}
            = \hat{\theta}_k
              + \alpha_k e_k \varphi_k
              + \beta \bigl(\hat{\theta}_k - \hat{\theta}_{k-1}\bigr).
     \end{equation}
    \EndFor
\end{algorithmic}
\end{algorithm}
The momentum term
$
\beta\left(\hat{\theta}_k-\hat{\theta}_{k-1}\right)
$ in Algorithm \ref{alg:mnlms} incorporates the previous update direction. This may allow the algorithm to smooth parameter updates, accelerate convergence, and track time-varying parameters more effectively, which distinguishes MLMS from the classical LMS algorithm.
The regularization constant $\delta$ is chosen empirically for numerical stability and is not sensitive within a reasonable range.

The goal of Algorithm~\ref{alg:mnlms} is to track the time-varying parameter process
$\{\theta_k\}$. The tracking performance depends on the parameter-variation
process $\{\Delta_k\}$, defined by
\begin{equation}\label{parametererror}
  \Delta_{k}
  = \theta_{k}-\theta_{k-1},
\end{equation}
and assume that $\theta_{-1}=\theta_0 $.
Set
\[
\tilde{\theta}_{k} \triangleq \theta_{k}-\hat{\theta}_{k},
\qquad
d_k \triangleq \hat{\theta}_k-\hat{\theta}_{k-1},
\qquad
\eta_k \triangleq
\frac{v_{k+1}\varphi_k}{\delta+\|\varphi_k\|^2}.
\]
Substituting \eqref{timevarying1} and \eqref{parametererror} into \eqref{eq:hfwhif} and noticing the definition of $e_k$ yields the following coupled error recursion:
\begin{equation}\label{eq:error_recursion}
\begin{aligned}
\tilde{\theta}_{k+1}
&=
\left(
I-\frac{\mu\,\varphi_k\varphi_k^{\top}}
       {\delta+\|\varphi_k\|^2}
\right)
\tilde{\theta}_k
-\beta d_k
+\Delta_{k+1}
-\mu\eta_k,
\\
d_{k+1}
&=
\frac{\mu\,\varphi_k\varphi_k^{\top}}
     {\delta+\|\varphi_k\|^2}
\tilde{\theta}_k
+\beta d_k
+\mu\eta_k .
\end{aligned}
\end{equation}

Directly analyzing \eqref{eq:error_recursion} to obtain the tracking performance is challenging due to the coupling among the estimation terms. To facilitate the analysis of the parameter-estimation error, we stack the current error vector and the momentum increment so that the recursion can be rewritten in the following state space form. Define
\[
A_k = \frac{\mu\,\varphi_k\varphi_k^{\top}}
           {\delta+\|\varphi_k\|^2}
\]
and
\[
Z_k
\triangleq
\begin{pmatrix}
\tilde{\theta}_k\\[2pt]
d_k
\end{pmatrix}
=
\begin{pmatrix}
\tilde{\theta}_k\\[2pt]
\hat{\theta}_k-\hat{\theta}_{k-1}
\end{pmatrix}.
\]
Then \eqref{eq:error_recursion} becomes
\begin{align}
Z_{k+1}
&= \begin{pmatrix}
I-A_k & -\beta I\\[2pt]
A_k & \beta I
\end{pmatrix}
Z_k
+
\begin{pmatrix}
\Delta_{k+1}-\mu\eta_k\\[2pt]
\mu\eta_k
\end{pmatrix}\notag\\[2pt]
&=
\bigl(I_0-\bar{A}_k\bigr)Z_k
+
\begin{pmatrix}
\Delta_{k+1}-\mu\eta_k\\[2pt]
\mu\eta_k
\end{pmatrix},
\label{eq:error_equation}
\end{align}
where
\[
I_0 =
\begin{pmatrix}
I & \mathbf{0}\\[2pt]
\mathbf{0} & \mathbf{0}
\end{pmatrix},
\qquad
\bar{A}_k =
\begin{pmatrix}
A_k & \beta I\\[2pt]
-A_k & -\beta I
\end{pmatrix}.
\]

Equation~\eqref{eq:error_equation} provides a linear state-space representation of the
error dynamics, which will be used as the basis for the subsequent analysis. The properties of \eqref{eq:error_equation} are essentially
determined by its homogeneous equation
\begin{equation}\label{LALALAL}
Z_{k+1} = \bigl(I_0-\bar{A}_k\bigr)Z_k .
\end{equation}

\begin{rem}
It is worth noting that,  \eqref{LALALAL} is fundamentally different from that used for the analysis traditional LMS (see, e.g., \cite{guo1994stability, GuoLjungPriouret1997}), because $I_0$ is no longer the identity matrix, and the $\bar{A}_k$ is no longer a symmetric matrix.  This significant difference results in the key difference between the analysis of MLMS and LMS.
\end{rem}
\section{Main Results}\label{Results}
In this section, we establish the tracking performance and regret of the algorithm introduced in Section \ref{hello}. 
We now present the following theorem to characterize the exponential stability of \eqref{LALALAL}, which serves as an essential foundation for analyzing the parameter tracking performance.
\begin{thm}\label{lemma11}
Under Assumption~\ref{condition2}, for any integer $p \ge 1$, there exists a constant $M$ such that
\begin{equation}
\left\|
\prod_{\ell=i+1}^{k}\left(I_0-\bar A_\ell\right)
\right\|_{p}
\leq
M\lambda_p^{k-i},
\quad
\forall k\geq i\geq 0,
\label{eq:Thm4_new}
\end{equation}
where $\lambda_{p} = \bigl(1 - \tfrac{\alpha \mu}{8}\bigr)^{\frac{1}{ph}} \in (0,1)$,
the hyperparameter $\mu$ is chosen to satisfy
\begin{equation}\label{mucondition}
\begin{aligned}
0 < \mu &\le 
\min\Bigl\{
1,\left(h \sqrt{2\left(1+C_\beta^2\right)}\right)^{-1},\left(q C_K\right)^{-\frac{1}{2}},\\
& \qquad\quad \frac{1}{3 \alpha}, \frac{\alpha}{6 C_\beta^2}, \frac{\alpha}{6 C_K},\left(\frac{\alpha}{3 C_K^2}\right)^{\frac{1}{3}}, \frac{\alpha}{8 D_q}\Bigl\},
\end{aligned}
\end{equation}
where $q = p-\tfrac12$,  $C_K=\frac{h^2+1+3 C_\beta^2}{2}+2(e-2) h^2\left(1+C_\beta^2\right)$ and $D_q=q C_K+(e-2) q^2 C_K^2$ .

\end{thm}
\begin{rem}
The bound on $\mu$ is introduced to ensure the $L_p$-stability of the subsequent parameter estimation error equation \eqref{eq:error_equation}, which forms the basis for analyzing the tracking performance. As demonstrated in the proofs of the main theorems, the bound arises naturally from controlling the worst case effects of the noises and the parameter variations. Notably, the specific form given in Theorem~\ref{lemma11} provides a sufficient condition that facilitates a unified analysis. In practical implementations of the algorithm to a specific problem, however, the admissible range of $\mu$ may be larger and the step-size can be further  tuned based on empirical performance.
\end{rem}
\begin{rem}
    The tracking performance of the algorithm depends critically on the stochastic
properties of the processes $\{\varphi_k, \Delta_k, v_k\}$.
The homogeneous recursion \eqref{LALALAL} is handled by Theorem~\ref{lemma11}, so it
remains to study the effect of the nonhomogeneous term in \eqref{eq:error_equation}.
\end{rem}
Different assumptions on $\left\{\Delta_k, v_k\right\}$ lead to different tracking-error bounds. In the following, we consider two separate cases for $\left\{\Delta_k, v_k\right\}$.
\subsection{The Case of Bounded Noises and Parameter Variations}\label{First}
We first analyze the tracking performance in a
``worst-case'' scenario, where the parameter variations and disturbances are
assumed to be bounded only in an averaged sense.
\begin{assum}\label{assumpfirst}
    For some integer $p \ge 1$, $\|\tilde{\theta}_0\|_{2p} < \infty$, and
\[
\sigma \triangleq \sup_{k}\|v_k\|_{2p} < \infty,
\qquad
\nu \triangleq \sup_{k}\|\Delta_k\|_{2p} < \infty.
\]
Note that this assumption includes, as a special case, any deterministic
``unknown but bounded'' disturbances and parameter variations.
\end{assum}
\begin{thm}\label{theorem1}
Suppose that the conditions of Theorem~\ref{lemma11} and Assumption~\ref{assumpfirst} hold.
Then, for all $k \ge 1$,
\begin{equation}\label{hahahhaaaaa}
\bigl\|\tilde{\theta}_{k+1}\bigr\|_{p}
=
O\!\Bigl(
  \lambda_{2 p}^{k+1} \bigl\|\tilde{\theta}_0\bigr\|_{2p}
\Bigr)
\;+\;
O\!\Bigl(
  \tfrac{\nu}{\mu} + \sigma
\Bigr),
\end{equation}
where $\lambda_{2p}$ is given in Theorem~\ref{lemma11}. 
\end{thm}
 We remark that the first term on the right hand side of \eqref{hahahhaaaaa} tends to zero exponentially fast and that the 
asymptotical tracking error will be small provided that both the levels of the parameter variation
$\nu$ and the disturbance $\sigma$ are small. With \(0\le\beta\le C_\beta\mu\), the effect of $\beta$ is absorbed into the constants in the bound. In the upper bound on $\mu$ employed in Theorem~\ref{theorem1}, $q$ is taken as $q = 2p - \tfrac{1}{2}$ instead of
$q = p - \tfrac{1}{2}$, so as to enable the application of the Schwarz inequality in 
establishing the tracking performance bound \eqref{hahahhaaaaa}. Details can be found in the proof of Theorem \ref{theorem1}.
\subsection{The Case of Zero Mean Random Noises and Parameter Variations}\label{Second}
We next analyze the tracking performance in the case of zero-mean random
parameter variations and disturbances, which are allowed to be general
correlated processes. To this end, we introduce the following class
of stochastic processes for $p \ge 1$:
\begin{equation}\label{wula}
\mathcal{M}_p
\triangleq
\Bigl\{
  \{x_k\} :
  \bigl\|\sum_{i=k+1}^{k+n} x_i\bigr\|_p
  \le c_p^x \sqrt{n},
  \ \forall k \ge 0,\ \forall n \ge 1
\Bigr\},
\end{equation} for $p \ge 1$ 
where $c_p^x$ is a constant depending only on $p$ and on the moment
of the process $\{x_k\}$. It is known that martingale-difference sequences,
zero-mean $\phi$-mixing and zero-mean $\alpha$-mixing sequences all belong
to $\mathcal{M}_p$ (see, e.g., \cite{1992GuoTracking}).
\begin{assum}\label{assumpsecond}
    For some integer $p \ge 1$, $\|\tilde{\theta}_0\|_{2p} < \infty$, and
\[
\Bigl\{
  \tfrac{\varphi_k v_{k+1}}{\delta + \|\varphi_k\|^2}
\Bigr\}
\in \mathcal{M}_{2p},
\qquad
\{\Delta_k\} \in \mathcal{M}_{2p}.
\]
\end{assum}
\begin{thm}\label{theorem2}
Suppose that the conditions of Theorem~\ref{lemma11} and Assumption \ref{assumpsecond} hold. Then, for all $k \ge 1$,
\begin{equation}
\bigl\|\tilde{\theta}_{k+1}\bigr\|_{p}
=
O\!\Bigl(
  \lambda_{2p}^{k+1}
  \bigl\|\tilde{\theta}_0\bigr\|_{2p}
\Bigr)
\;+\;
O\left(
c_{2p}^{\Delta}\mu^{-\frac12}
+
c_{2p}^{v}\mu^{\frac12}
\right), 
\end{equation}
where $\lambda_{2p}$ is given in Theorem~\ref{lemma11}, $c_{2p}^{\Delta}$ and $c_{2p}^v$ are the constants defined in \eqref{wula}, 
for the random processes $\{\Delta_k\}$ and 
$\left\{\tfrac{\varphi_k v_{k+1}}{\delta + \|\varphi_k\|^2}\right\}$, respectively.
\end{thm}
We remark that the upper bound in Theorem~\ref{theorem2} significantly
improves the crude bound of Theorem~\ref{theorem1}, and it clearly exhibits
the familiar tradeoff between noise sensitivity and tracking ability. The explanation of the influence of $\beta$ follows the same reasoning as in Theorem~\ref{theorem1}. In the upper bound on $\mu$ employed in Theorem~\ref{theorem2}, $q$ is also taken as $q = 2p - \tfrac{1}{2}$ for the same reason as in Theorem \ref{theorem1}.

\subsection{Prediction Analysis }
In Subsections~\ref{First} and \ref{Second}, we analyzed the parameter tracking performance under the conditional excitation condition given in Assumption~\ref{condition2}.
In this subsection, we analyze the output prediction performance of the MLMS algorithm without imposing any excitation conditions. To proceed, we require the following assumptions and definition.

\begin{assum}\label{qafbjk} 
The noise $\{v_{k+1}, \mathcal{F}_k\}$ is a martingale difference sequence. Moreover, there exist constants $\sigma_v>0$ such that
$
\mathbb{E}\left[v_{k+1}^2 \mid \mathcal{F}_k\right]\le \sigma_v^2,\quad a.s.,
$
and
$
\sup_{k\ge 0}\mathbb{E}\left[|v_{k+1}|^r \mid \mathcal{F}_k\right]<\infty,\quad a.s.,
$
for some $r\ge 4$.
\end{assum}
\begin{assum}\label{qqqqq}
The parameter vector \(\theta_k\) belongs to a known convex compact set
\[
\mathcal{D}=\left\{x=\left(x_1, \cdots, x_{m}\right)^{\top} \in \mathbb{R}^{m},
\left|x_i\right| \leqslant L, 1 \leqslant i \leqslant m\right\}.
\]
Moreover, the parameter process $\{\theta_k\}$ is slowly time-varying in the following Cesàro-average sense, i.e.,
\begin{equation} \label{bountheta}
\limsup _{n\to\infty} \frac{1}{n}\sum_{k=1}^{n} \|\theta_k - \theta_{k-1}\|
\leq \xi , \quad \text{a.s.},
\end{equation}
where $\xi >0$ is a suitably small positive constant.
The stochastic regressor sequence $\left\{\varphi_k, \mathcal{F}_k,k \geq 0\right\}$ is uniformly bounded, i.e., there exists a constant $C$ such that
$
\sup _{k\ge 0}\left\|\varphi_k\right\| \leq C .
$
\end{assum}

We remark that \eqref{bountheta} is more general than the condition
$
\left\|\theta_k - \theta_{k-1}\right\| \leq \xi, \ \forall k \geq 1
$, since it allows occasionally large jumps in the parameter process.

Furthermore, we consider simple modifications to Algorithm~\ref{alg:mnlms}, such as incorporating a projection operator or adding a $\sigma$-modification term \citep{MEYN1990149, ioannou1996robust}.
This modification ensures bounded parameter estimates, which not only maintains stability under perturbations but also allows us to derive a prediction bound without relying on any excitation assumptions, resulting in a more realistic theoretical guarantee.

Here, we focus on the projection-based update, while noting that the $\sigma$-modification may be analyzed in a similar manner as in \cite{MEYN1990149, ioannou1996robust}, and we refer the reader to these works for further details.

We first introduce the definition of projection.
\begin{defn}\citep{cheney2001analysis}\label{projection}
The projection operator $\Pi_{\mathcal{D}}(\cdot)$ is defined as
\begin{equation}
   \pi_{\mathcal{D}}(x)=\underset{y \in \mathcal{D}}{\arg \min }\|x-y\|, \quad \forall x \in \mathbb{R}^m ,
\end{equation}
where $\mathcal{D}$ is the compact convex set introduced in Assumption \ref{qqqqq}.
\end{defn}
Now, we introduce the specific modification to Algorithm~\ref{alg:mnlms}. This modification applies a projection to the parameter estimates, while all other updates remain unchanged.
\begin{equation}\label{projection12412}
\hat{\theta}_{k+1}
= \pi_{\mathcal D}\left[\hat{\theta}_k
  + \alpha_k e_k \varphi_k
  + \beta\bigl(\hat{\theta}_k-\hat{\theta}_{k-1}\bigr)\right].
\end{equation}
We note that if the parameter estimates remain within
$\mathcal{D}$ throughout the iterations, the projection operator has no effect, and the method simplifies to the original Algorithm~\ref{alg:mnlms}. We now establish a theoretical upper bound on the output prediction of the projection MLMS algorithm.
\begin{thm} \label{proposition1}
Under Assumptions~\ref{qafbjk} and~\ref{qqqqq}, we have the following averaged prediction error of the projection Algorithm \eqref{projection12412}:
\begin{equation}\label{bound-error-upper}
\begin{aligned}
& \limsup _{n\to\infty} \frac{1}{n}\sum_{k=1}^{n}  \bigl(y_{k+1}-\hat y_{k+1}\bigr)^2 \\
&\leq \sigma_v^2+O\left(\mu \sigma_v^2+\frac{\beta}{\mu}+\beta \sigma_v+\frac{\xi}{\mu}\right),\quad \text{a.s.}
\end{aligned}
\end{equation}
\end{thm}
From Theorem~\ref{proposition1}, one can see that the
upper bound of the averaged prediction error depends explicitly on
both the step-size $\mu$ and the momentum coefficient $\beta$.
In the special case of constant parameters, i.e., $\xi=0$, and further choose$\beta=o(\mu)$, the excess term in the above
upper bound vanishes as $\mu\to0$. Hence,
\[
\limsup _{n\to\infty}
\frac{1}{n}
\sum_{k=1}^{n}
\bigl(y_{k+1}-\hat y_{k+1}\bigr)^2
\leq
\sigma_v^2+o(1),
\quad \text{a.s.}
\]
Thus, in this small-step-size regime with asymptotically
negligible momentum, the upper bound approaches the noise level
$\sigma_v^2$.
\section{Formal Linear Representation}\label{formal}
In this section, we consider a basic question: \emph{How do LMS-type methods work beyond linear models?} The LMS algorithm was originally devised for linear, time-varying identification, yet in practice it often performs well on genuinely nonlinear problems.  Consider the following nonlinear observation model
\begin{equation}\label{bobibobi}
y_{k+1} = f(\Theta_k,\phi_k) + v_{k+1}.
\end{equation}
The key observation is that this model admits an exact linear time-varying representation once a path-averaged Jacobian with respect to the regressor is introduced.

\noindent\textbf{Setup and definitions.}
Let \(f:\mathbb{R}^{d_\theta}\times\mathbb{R}^{d_\phi}\to\mathbb{R}\) be differentiable. For a latent parameter process \(\{\Theta_k\}\subset\mathbb{R}^{d_\theta}\) and regressors \(\{\phi_k\}\subset\mathbb{R}^{d_\phi}\), fix any \emph{predictable} reference sequence \(\{\bar\phi_k\}\subset\mathbb{R}^{d_\phi}\) (for instance, \(\bar\phi_k\equiv 0\), a running mean, or an \(\mathcal{F}_{k}\)-measurable baseline). Define the path-averaged Jacobian along the segment from \(\bar\phi\) to \(\phi\) by
\[
A(\Theta;\phi\,;\,\bar\phi)
\;= \;
\int_{0}^{1} \nabla_{\phi} f\bigl(\Theta,\,\bar\phi + t(\phi-\bar\phi)\bigr)\,dt
\;\in\; \mathbb{R}^{d_\phi}.
\]
By the fundamental theorem of calculus applied to the curve
\(t\mapsto f\bigl(\Theta,\bar\phi+t(\phi-\bar\phi)\bigr)\),
we obtain the exact identity
\begin{equation}\label{eq:FLR-exact}
f(\Theta,\phi)
\;=\;
f(\Theta,\bar\phi)
\;+\;
A(\Theta;\phi\,;\,\bar\phi)^\top\,(\phi-\bar\phi),
\end{equation}
which is a global equality, not a local approximation. Now, we give the exact time-varying linear representation. 

\medskip
\noindent\textbf{Formal Linear Representation (FLR).}
Using \eqref{eq:FLR-exact}, the observation model~\eqref{bobibobi} can be written as a linear, time-varying regression:
\begin{align}
y_{k+1}
&= f(\Theta_k,\phi_k) + v_{k+1} \nonumber\\
&= f(\Theta_k,\bar\phi_k)
   + A(\Theta_k;\phi_k\,;\,\bar\phi_k)^\top\,(\phi_k-\bar\phi_k)
   + v_{k+1} \nonumber \\
&= \underbrace{\begin{bmatrix} 1 \\[2pt] \phi_k-\bar\phi_k \end{bmatrix}^{\top}}_{\displaystyle \varphi_k\in\mathbb{R}^{1+d_\phi}}
   \;
   \underbrace{\begin{bmatrix}
   f(\Theta_k,\bar\phi_k)\\[2pt]
   A(\Theta_k;\phi_k\,;\,\bar\phi_k)
   \end{bmatrix}}_{\displaystyle \theta_k\in\mathbb{R}^{1+d_\phi}}
   \;+\; v_{k+1} \nonumber\\
&= \varphi_k^\top \theta_k + v_{k+1}. 
\end{align}
\begin{rem}
The nonlinear model \eqref{bobibobi} can be rewritten exactly as the ``linear'' time-varying regression form \eqref{timevarying1}, where ``linear'' only means that the unknown parameters $\theta_k$ enter into the model linearly and does not mean that the system is linear. This ``linear'' representation is precisely the setting for which LMS-type algorithms were originally developed. This reformulation may explain why such algorithms are capable of handling a broad class of nonlinear systems.
\end{rem}
\begin{rem}
The formulation above represents one exact, parameterized ``linear'' approach. Besides, there are also other ways to construct ``linear'' representation. For example, the function \(f(\Theta_k,\phi_k)\) can be approximated by maximizing a bilinear functional \citep{song2025remarks}, or by applying a Mercer expansion under continuous kernels \citep{jeong2024extending}. Another classical method is based on the Koopman operator (see, e.g., \cite{Koopman1931}), which embeds nonlinear dynamics into an infinite-dimensional linear operator to obtain linear evolution in lifted feature spaces.
Since these alternatives are not the main focus of this paper, we leave a detailed discussion to future work. 
\end{rem}
\section{Experiments}\label{experiments}
While the MLMS algorithm with an unnormalized LMS update (i.e., $\alpha_k$ in Algorithm~\ref{alg:mnlms} is a constant $\mu$ in their setting) has been extensively studied, its normalized counterpart, where $\alpha_k = \frac{\mu}{\delta + \|\varphi_k\|^2}$ in our formulation, has received far less empirical investigation. To this end, we examine the practical behavior of Algorithm~\ref{alg:mnlms} and compare it with several commonly used stochastic optimization methods, including the classical unnormalized LMS and its momentum variant (represented here by SGD and SGD-Momentum, respectively), as well as the normalized LMS (represented here as LMS).
\subsection{Synthetic Data with Jumping Parameters}
Consider the following linear stochastic system:
\begin{equation}\label{eq:jump-system}
\left\{
\begin{aligned}
&\varphi_{k+1} = A \varphi_k + v_k,\\[1mm]
&\theta_k = \theta_{k-1} + \Delta_k,\\[1mm]
&y_k = \varphi_k^\top \theta_k + \varepsilon_k,
\end{aligned}
\right.
\end{equation}
where $A=\mathrm{diag}[0.6,\,0.7,\,0.9,\,0.2,\,0.5,\,0.3]$ denotes the diagonal matrix, $\varphi_k \in \mathbb{R}^6$ with $\varphi_0=\mathbf{0}$. The state noise $v_k \sim \mathcal{N}(0,4I_6)$. The parameter process $\left\{\theta_k\right\}$ is initialized at $\theta_0 \sim \mathcal{N}\left(0, I_6\right)$ and remains constant between abrupt change points. Every $100$ iterations, the parameter vector undergoes a sudden jump:\[
\Delta_k =
\begin{cases}
0, & k \neq  0 \pmod{100},\\
0.5 \, \zeta_k, & k = 0 \pmod{100},
\end{cases}
\quad \zeta_k \sim \mathcal{U}([-1,1]^6),
\]
where $\mathcal{U}([-1,1]^6)$ denotes the uniform distribution over the
six-dimensional hypercube $[-1,1]^6$.
This construction yields a nonstationary process with persistent parameter changes. The observation noise $\{\varepsilon_k\}$ is i.i.d as $\varepsilon_k \sim \mathcal{N}(0, 0.1^2)$.

\textbf{Evaluation metrics.}  
Since the tracking error may span several orders of magnitude, We measure performance using the commonly used tracking MSE in dB, defined as
\begin{equation}
\mathrm{MSE}_{\mathrm{dB}}(k)
= 10\log_{10}\!\left( \|\hat{\theta}_k - \theta_k\|^2 + 10^{-12} \right),
\end{equation}
where the small constant $10^{-12}$ is added for numerical stability.

\textbf{Performance Comparison.} To demonstrate the advantages of the MLMS algorithm, we compare its  tracking performance against SGD, SGD-Momentum and LMS on a dataset $\{\varphi_k, y_{k}\}_{k=1}^{T=500}$ generated from the system~\eqref{eq:jump-system}. All algorithms are implemented exactly following their standard online-update forms. The hyperparameter configuration is: 
SGD and SGD-Momentum use learning rate $10^{-3}$, with momentum $0.1$ for the latter; LMS uses a step-size of $0.1$ and a regularization term of $0.1$, while MLMS employs the same step-size and regularization term, together with a momentum coefficient of $\beta=0.099$.

\begin{figure}[htbp]
\begin{center}
\includegraphics[width=1\linewidth]{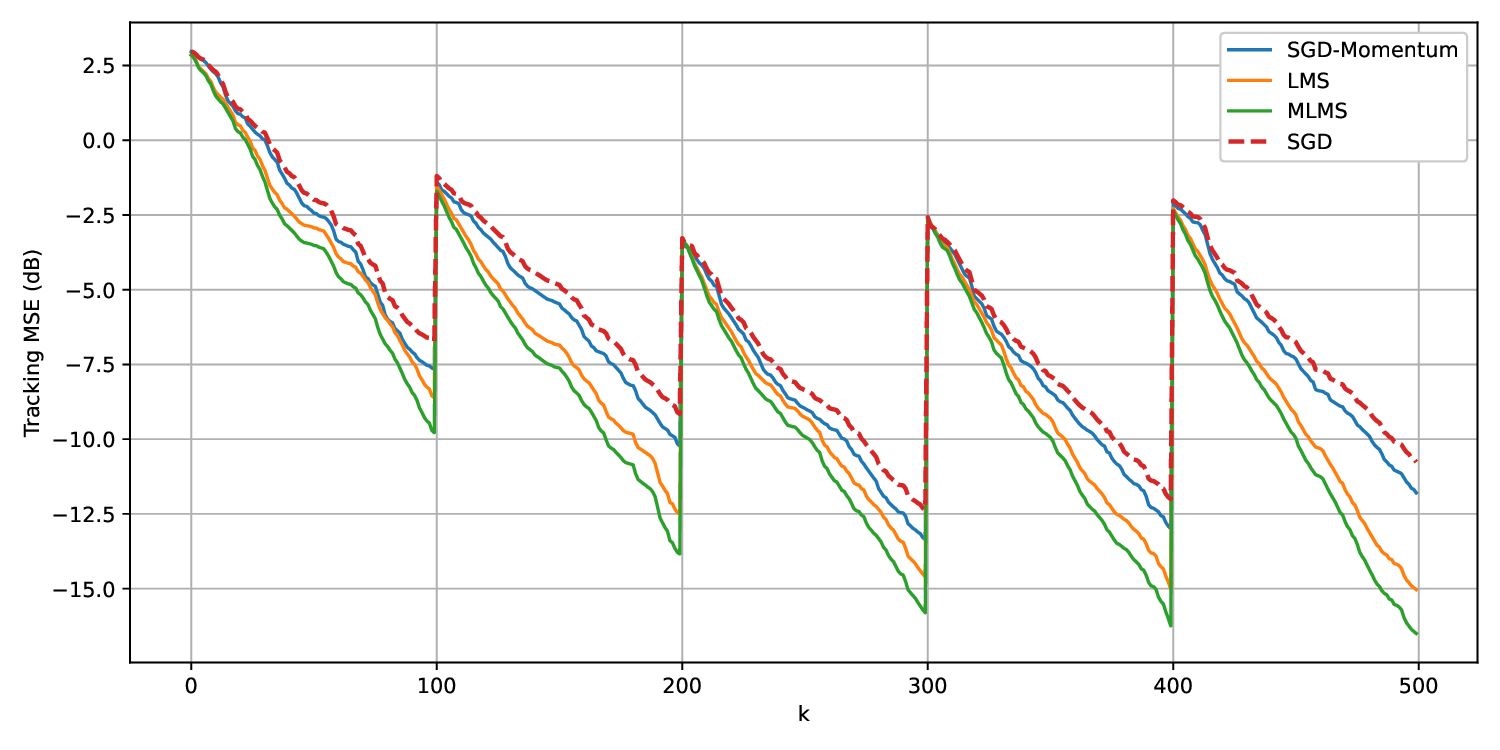}
\caption{Tracking MSE (in dB) over 10 independent trials for the time-varying regression task.}
\label{fig:HASU-column}
\end{center}
\end{figure}

As shown in Fig~\ref{fig:HASU-column}, the frequent abrupt changes in $\theta_k$ create a nonstationary environment. Algorithms equipped with momentum respond more rapidly to these jumps, exhibiting noticeably faster re-convergence after each change point. Among all methods, the MLMS algorithm achieves the lowest parameter tracking MSE, highlighting its superior adaptability in scenarios with abrupt changes in the underlying parameter.

\subsection{Application to Speech Enhancement}
In real-world acoustic environments, speech signals are frequently contaminated by nonstationary background noise arising from evolving and unpredictable sources (see, e.g., \cite{loizou2007speech}). Consequently, online adaptive filtering has become a fundamental tool for real-time speech enhancement, as it can track dynamic signal. To evaluate the MLMS algorithm in such streaming conditions, we conduct experiments on the NOIZEUS corpus\footnote{\url{https://ecs.utdallas.edu/loizou/speech/noizeus/}} with airport noise.

\textbf{Evaluation metrics.} Performance is evaluated using the commonly adopted signal-to-noise ratio (SNR), computed online during streaming.
Let $n_k$ denote the noisy observation, $c_k$ the clean reference, and $y_k$ the filter output at time $k$.
We compute the input and output SNRs as
\[
\mathrm{SNR}_{\mathrm{out}} = 10\log_{10}\!\frac{\sum_k c_k^2}{\sum_k (y_k - c_k)^2}, 
\]
\[
\mathrm{SNR}_{\mathrm{in}} = 10\log_{10}\!\frac{\sum_k c_k^2}{\sum_k (n_k - c_k)^2}.
\]
The reported SNR improvement is then
$ 
\Delta\mathrm{SNR}
= \mathrm{SNR}_{\mathrm{out}} - \mathrm{SNR}_{\mathrm{in}}.
$ 

We evaluate six adaptive filtering algorithms under real-world airport noise at
three input SNR levels (5, 10, and 15 dB): SGD, SGD-Momentum,
RLS, GNGD (generalized normalized gradient descent, see \cite{mandic2004generalized}), LMS and MLMS. For each noise
condition, 30 utterances from the NOIZEUS corpus are processed independently in
a streaming, sample-by-sample fashion. All algorithms use 50-tap adaptive
filters, and filter states are
reinitialized between utterances. The hyperparameter settings are as follows: SGD uses a
step-size of $\mu = 0.1$, while SGD-Momentum applies the same step-size together
with a momentum coefficient of $\beta = 0.2$; LMS uses $\mu = 0.25$ and $\delta=10^{-12}$, while MLMS employs
the same step-size and regularization term with momentum coefficient $\beta = 0.15$. For RLS, we
use a forgetting factor of $\lambda = 0.999$. The GNGD filter is configured with initial values $\rho = 0.01$, and step-size $\mu = 0.1$.

\begin{table}[htbp]
\centering
\caption{SNR improvement (dB) under different noise levels. Higher is better.}
\begin{tabular}{lccc}
\toprule
\textbf{Filter} & \textbf{5 dB} & \textbf{10 dB} & \textbf{15 dB} \\
\midrule
SGD             & $7.72 \pm 1.47$ & $6.39 \pm 3.38$ & $4.94 \pm 7.38$ \\
SGD-M    & $7.74 \pm 1.61$ & $6.42 \pm 4.09$ & $4.91 \pm 10.58$ \\
RLS             & $3.92 \pm 2.75$ & $2.84 \pm 6.70$ & $3.08 \pm 15.31$ \\
GNGD            & $6.44 \pm 1.32$ & $5.14 \pm 3.00$ & $3.76 \pm 7.43$ \\
LMS             & $7.86 \pm 1.52$ & $6.79 \pm 3.29$ & $5.69 \pm 8.63$ \\
\textbf{MLMS}   & $\mathbf{7.91 \pm 1.57}$ & $\mathbf{6.86 \pm 3.36}$ & $\mathbf{5.79 \pm 8.78}$ \\
\bottomrule
\end{tabular}
\label{tab:snr_comparison}
\end{table}
Building on this setup, we aggregate the SNR improvements across the 30 utterances at each noise level and report the results as the mean ± standard deviation. As shown in Table~\ref{tab:snr_comparison}, MLMS consistently achieves the highest SNR gains across all noise conditions, demonstrating superior robustness in this realistic adaptive noise-suppression scenario.
\begin{figure}[htbp]
\begin{center}
\includegraphics[width=1\linewidth]{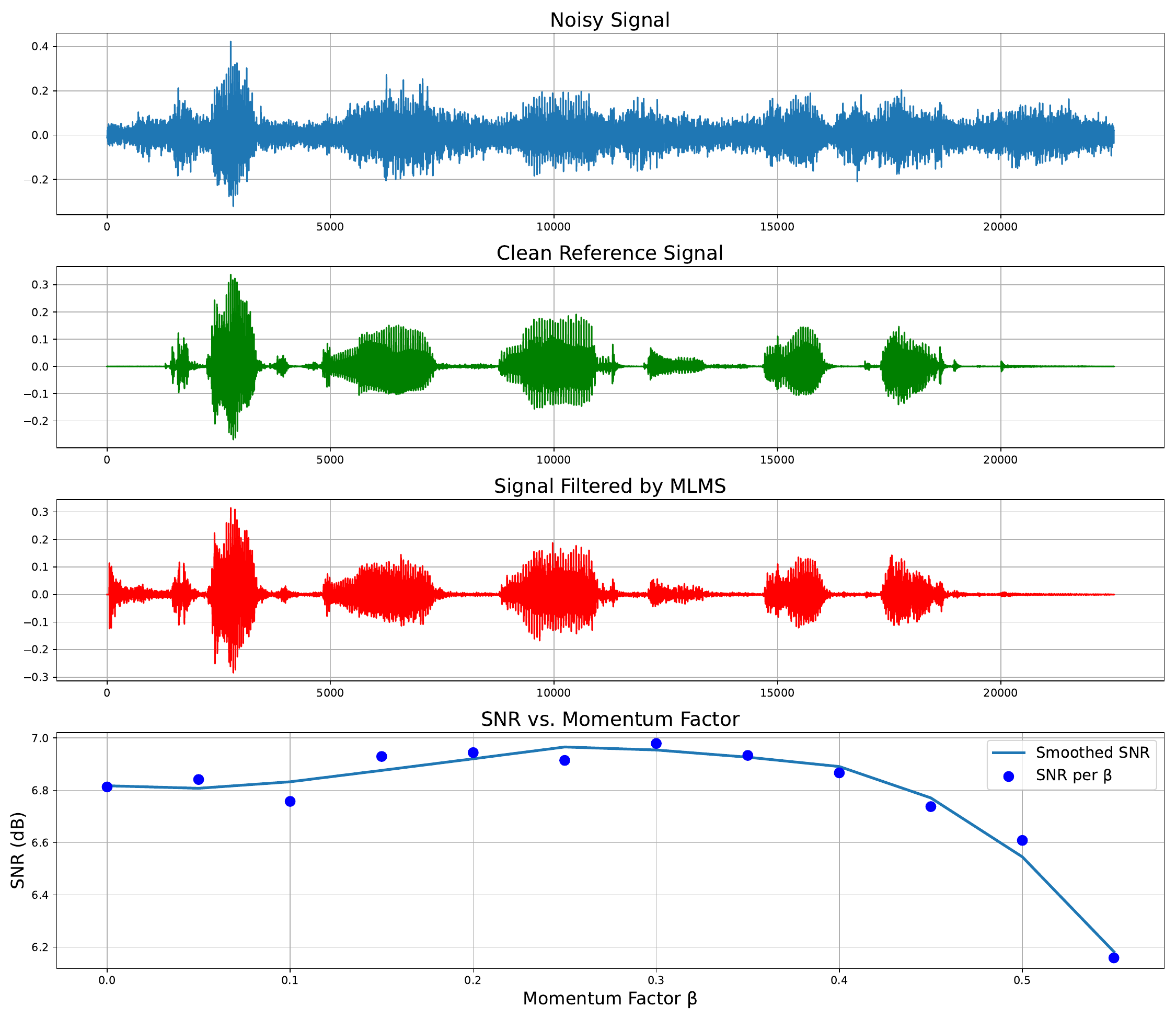}
\caption{Adaptive Noise Cancellation}
\label{fig:double-column}
\end{center}
\end{figure}

Next, we focus on a challenging 0 dB condition, where the clean speech and background noise have comparable magnitudes, making adaptive filtering
  particularly difficult due to the persistently low instantaneous SNR. The first three subplots of Fig.~\ref{fig:double-column} illustrate the
  time-domain waveforms of the noisy input, the clean reference, and the MLMS output. In this experiment, the main parameters are set to $\mu =
  0.35$, $\beta = 0.3$, and $\epsilon = 10^{-3}$. Even under this adverse scenario, the filtered waveform closely follows speech onsets and
  noticeably suppresses rapid fluctuations relative to the noisy signal.
  The fourth subplot of Fig.~\ref{fig:double-column} shows the output SNR as a function of the momentum factor $\beta$, averaged over 30
  utterances. The SNR first improves as $\beta$ increases and then degrades when excessive momentum is introduced, indicating that a moderate
  momentum term can accelerate adaptation, whereas overly large momentum may reduce tracking ability in nonstationary environment.

\section{Conclusion}\label{conclusion}
The MLMS algorithm has been widely acknowledged to have better convergence performance than the classical LMS algorithm in many practical situations, but has rarely been explored in theory with nonstationary datasets. In view of this, we have provided a theoretical foundation of the MLMS algorithm by establishing rigorous parameter tracking guarantees under the datasets characterized by the conditional excitation condition of \cite{guo2002estimating}, which represents a substantive advance over the widely adopted i.i.d. or stationary signal assumptions in the previous related investigations, and more importantly, retaining applicability to stochastic systems with feedback control where the i.i.d assumptions do not hold. In addition, we have derived prediction guarantees that require no excitation conditions at all, thereby significantly broadening the range of scenarios in which reliable performance can be ensured. By overcoming the analytical barrier that arises when the stability analysis transitions from a first-order random vector recursion in LMS to a second-order one in MLMS, this work establishes a practical and tractable theoretical framework for handling products of more general nonstationary matrices. Building on this foundation, the MLMS framework appears promising for learning or tracking of general nonlinear tine-varying systems, paving a way for more powerful momentum-enhanced adaptive algorithms in complex real-world.
\bibliographystyle{apalike2}       
\bibliography{main}
\end{document}